\documentclass{article}
\usepackage{spconf,amsmath,graphicx,booktabs,hyperref}
\hypersetup{hypertexnames=false}
\usepackage{array}
\usepackage{float}
\usepackage{placeins}
\usepackage{xspace}
\usepackage{multicol}
\usepackage[table]{xcolor}
\definecolor{tabbest}{RGB}{242,156,156}
\definecolor{tabsecond}{RGB}{244,200,150}
\definecolor{tabthird}{RGB}{245,232,180}
\definecolor{tabmem}{RGB}{224,224,224}
\newcommand{\best}[1]{\cellcolor{tabbest}#1}
\newcommand{\second}[1]{\cellcolor{tabsecond}#1}
\newcommand{\third}[1]{\cellcolor{tabthird}#1}
\newcommand{\membest}[1]{\cellcolor{tabmem}\textbf{#1}}

\newcommand{\ours}{AdaTex4D\xspace}

\title{AdaTex4D: Adaptive Texture Capacity Allocation for 4D Gaussian Splatting}
\name{De Jiang$^{1}$, Peiqiang Wang$^{2}$, Kehong Yuan$^{3}$, Shaohua Ma$^{4,*}$\thanks{\raggedright $^{*}$Corresponding author: Shaohua Ma (\texttt{ma.shaohua@sz.tsinghua.edu.cn}).}}
\address{$^{1,2,3,4}$ Tsinghua Shenzhen International Graduate School, Tsinghua University, Shenzhen, China}

\begin{document}
\raggedbottom
\ninept
\maketitle

\begin{abstract}
Textured Gaussians improve local appearance capacity, but assigning the same texture resolution to every primitive wastes storage on low-detail or weakly visible regions. We introduce \ours, an adaptive texture-capacity module for deformation-based 4D Gaussian Splatting. We first train a standard 4DGS model, then attach one packed local RGBA texture plane to each Gaussian and jointly fine-tune the 4DGS and texture parameters. The two texture axes grow independently according to visibility-normalized screen-space gradients and deformed local scales. Experiments on N3DV and PanopticSports show that \ours reduces texture storage by more than half while preserving reconstruction quality. Under fixed memory budgets, adaptive allocation also improves quality over uniform texture assignment and reduces overall model and peak memory. These results show that dynamic, anisotropic texture allocation provides a more efficient way to distribute local appearance capacity in 4D Gaussian representations.
\end{abstract}


\section{Introduction}
\label{sec:intro}

Neural radiance fields established high-quality novel-view synthesis through continuous volumetric functions \cite{mildenhall2020nerf}, while 3D Gaussian Splatting (3DGS) replaced expensive ray sampling with an explicit differentiable rasterizer \cite{kerbl2023gaussians}. Dynamic Gaussian methods model time-varying scenes through deformation, motion control, or frame-wise updates \cite{wu2024four,yang2024deformable,zhang2025mega,huang2024scgs,sun20243dgstream,lin2024gaussianflow}. These advances improve dynamic geometry and rendering efficiency, but each Gaussian still offers limited spatial variation inside its projected footprint. Per-primitive texturing increases local appearance capacity \cite{chao2025textured,rong2025gstex}.

This added capacity creates a new allocation problem. Uniform textured Gaussians assign the same square texture to every primitive, although projected footprint, visibility, motion, and local image detail vary substantially. Large textures are therefore also paid for by small, occluded, or visually simple Gaussians, while detail-critical regions are not distinguished from easy regions. Increasing the global texture resolution can improve local detail, but its memory cost grows for every primitive at once. Existing compact representations and compression methods reduce static Gaussian redundancy \cite{lu2024scaffold,lee2024compact,niedermayr2024compressed} or dynamic storage \cite{katsumata2024compact,hu20254dgc,kwak2025modec,zhang2025mega}. The key question here is where to place texture capacity.

\begin{figure}[t]
\centering
\makebox[\columnwidth][c]{\includegraphics[width=1.08\columnwidth,trim=4bp 0bp 2bp 0bp,clip]{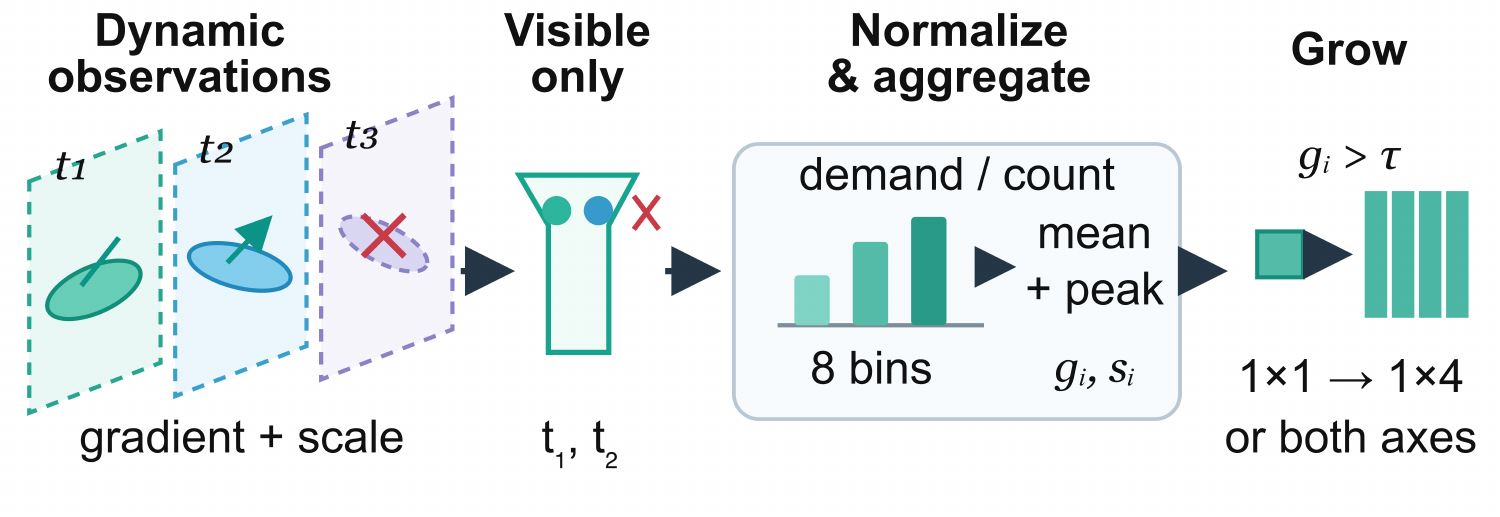}}
\caption{\textbf{Dynamic evidence aggregation in AdaTex4D.} Only visible dynamic observations contribute to each Gaussian's normalized gradient demand and deformed-scale statistics; these aggregated signals determine whether local texture capacity grows along one or both axes.}
\label{fig:dynamic-evidence}
\end{figure}

\begin{figure*}[t]
\centering
\includegraphics[width=0.90\textwidth,trim=10bp 0bp 10bp 0bp,clip]{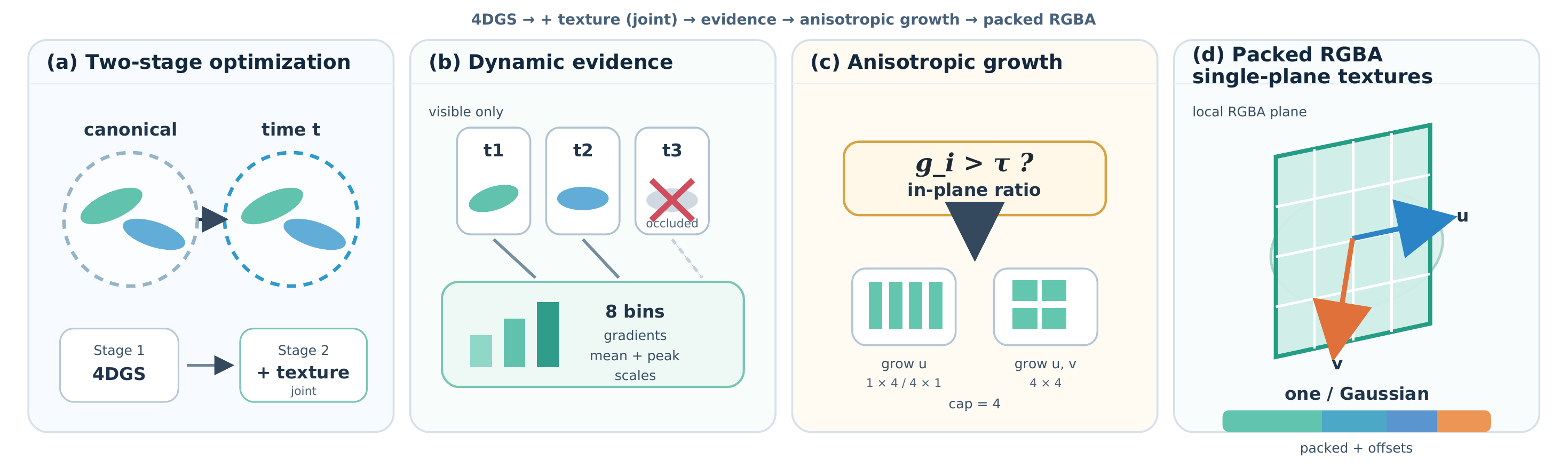}
\caption{\textbf{Overview of AdaTex4D.} Stage 1 trains 4DGS; Stage 2 jointly optimizes 4DGS and single-plane textures, with dynamic evidence driving anisotropic growth.}
\label{fig:overview}
\end{figure*}

\begin{figure*}[t]
\centering
\includegraphics[width=0.98\textwidth]{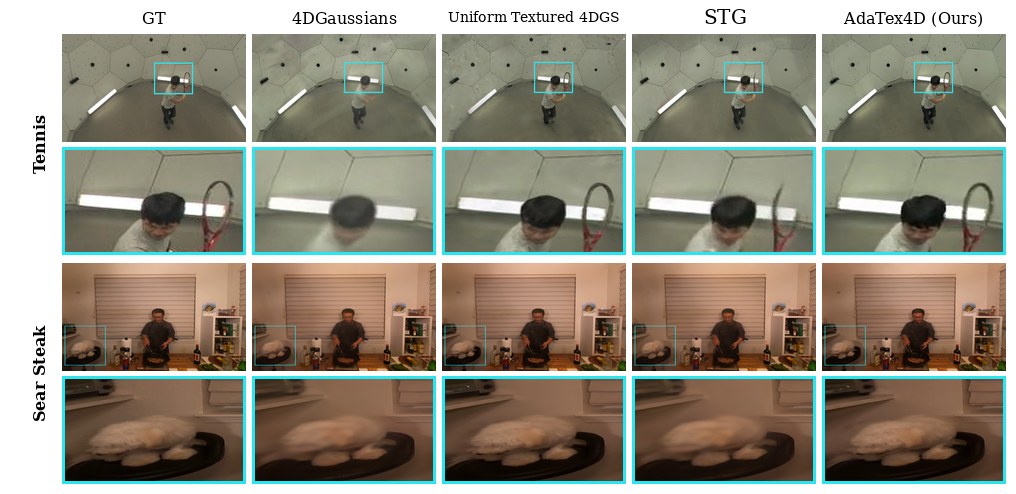}
\caption{\textbf{Qualitative comparison.} Full frames and enlarged ROIs for Tennis (top) and Sear Steak (bottom).}
\label{fig:qualitative}
\end{figure*}

Static A2TG grows textures from image-space gradients and Gaussian anisotropy \cite{hsu2026a2tg}. In a dynamic scene, a Gaussian may be visible in only part of the sequence, change scale after deformation, or carry fine detail for a short interval. We therefore aggregate visible observations over time and retain both their average and peak demand.

The allocation score uses visible observations, and deformed in-plane scales determine the growth direction. During second-stage optimization, Gaussian topology and primitive count stay fixed while the 4DGS parameters co-adapt with the texture branch.

\ours first trains a deformation-based 4DGS model. It then attaches one packed RGBA texture plane per Gaussian and jointly optimizes 4DGS and texture parameters with fixed Gaussian topology. Each adaptive plane starts at $1\!\times\!1$ and grows anisotropically to at most $4\!\times\!4$. Visibility-normalized screen-space gradients, temporal peak demand, and deformed local scales guide this growth.

Across 12 N3DV and PanopticSports scenes, the resulting allocation preserves reconstruction quality with substantially less texture state. Fixed-memory and fixed-Gaussian-count comparisons quantify the effects on reconstruction quality and storage.

We measure packed texture storage, inference-model size, peak GPU memory, training time, and FPS.

Our contributions are:
\begin{enumerate}
\setlength{\itemsep}{0pt}
\setlength{\parsep}{0pt}
\setlength{\topsep}{1pt}
\item We extend anisotropic texture allocation to deforming 4D Gaussians using visibility-aware temporal evidence and deformed local scales.
\item At 80 MiB, \ours improves PSNR by 0.20 and 0.18 dB over Uniform Textured 4DGS on N3DV and PanopticSports, respectively.
\item Across 12 scenes, \ours uses 47.3\% of the uniform texture bytes while preserving reconstruction quality.
\end{enumerate}

\section{Method}
\label{sec:method}

\subsection{Dynamic Gaussian parameterization}

After the first-stage 4DGS training, at time $t$ the model provides the center $\boldsymbol{\mu}_i^t$, rotation $\mathbf{R}_i^t$, scale $\mathbf{s}_i^t$, opacity $\alpha_i$, and SH color $\mathbf{c}_i^{\rm SH}$ for Gaussian $i$. In the second stage, these 4DGS parameters and the deformation field remain trainable and are jointly fine-tuned with the texture branch; only the Gaussian topology and primitive count are kept fixed. Following Textured Gaussians~\cite{chao2025textured}, each Gaussian carries a single local RGBA texture $\mathbf{T}_i$. Its texture plane is spanned by the two principal axes with the largest canonical scales; the remaining principal axis is the plane normal. The selected axis identities are fixed after initialization and are transported by the time-dependent Gaussian rotation, avoiding plane switching as the Gaussian deforms.

For a camera ray $\mathbf{r}(\lambda)=\mathbf{o}+\lambda\mathbf{d}$, let $p_i,q_i$ denote the two in-plane axis indices and $n_i$ the normal-axis index. The deformed plane normal is $\mathbf{n}_i^t=\mathbf{R}_i^t\mathbf{e}_{n_i}$. The screen-to-plane homography used by the renderer is equivalent to intersecting the ray with the plane through $\boldsymbol{\mu}_i^t$,
\begin{equation}
\lambda_i=\frac{(\mathbf{n}_i^t)^{\top}(\boldsymbol{\mu}_i^t-\mathbf{o})}{(\mathbf{n}_i^t)^{\top}\mathbf{d}},\qquad
\mathbf{x}_i^t=\mathbf{o}+\lambda_i\mathbf{d},
\label{eq:intersection}
\end{equation}
and map the intersection to normalized local texture coordinates
\begin{equation}
 u_i=\frac{\mathbf{e}_{p_i}^{\top}(\mathbf{R}_i^t)^{\top}(\mathbf{x}_i^t-\boldsymbol{\mu}_i^t)}{s_{i,p_i}^t},\qquad
 v_i=\frac{\mathbf{e}_{q_i}^{\top}(\mathbf{R}_i^t)^{\top}(\mathbf{x}_i^t-\boldsymbol{\mu}_i^t)}{s_{i,q_i}^t}.
\label{eq:uv}
\end{equation}
Coordinates $(u_i,v_i)$ are clamped to $[-1,1]^2$, mapped to the current texture grid, and bilinearly sampled from packed texels. Let $(\mathbf{r}_i,a_i^{\rm tex})$ be the sampled raw RGBA parameters. The RGB term adds a bounded residual to the jointly optimized SH appearance, while the alpha parameter multiplies the Gaussian opacity by a positive factor. Rendering is
\begin{equation}
\begin{aligned}
\widehat{\mathbf{I}}&=\sum_i w_i\left[\mathbf{c}_i^{\rm SH}+0.1\tanh(\mathbf{r}_i)\right]+T_{\rm final}\mathbf{b},\\
w_i&=\widetilde{\alpha}_i\prod_{j<i}(1-\widetilde{\alpha}_j),
\end{aligned}
\label{eq:render}
\end{equation}
where $\widetilde{\alpha}_i=\min\{0.99,\alpha_i G_i\exp[0.1\tanh(a_i^{\rm tex})]\}$, $G_i$ is the projected Gaussian falloff, $T_{\rm final}=\prod_i(1-\widetilde{\alpha}_i)$, and $\mathbf{b}$ is the background color. The renderer retains the 4DGS screen-space ellipse and front-to-back compositing. Each adaptive texture starts at $1\!\times\!1$; its two in-plane axes may grow through $\{1,2,4\}$, while the uniform control assigns a fixed $4\!\times\!4$ RGBA texture to every Gaussian.

\subsection{Visibility-normalized dynamic evidence}

A2TG uses image-space gradients to allocate texture capacity in static scenes \cite{hsu2026a2tg}. For visible occurrence $k$ of Gaussian $i$, we use the projected-center gradient magnitude
\begin{equation}
q_{i,k}=\bigl\|\nabla_{\mathbf{m}_{i,k}}\mathcal{L}_k\bigr\|_2
\label{eq:grad-demand}
\end{equation}
as its texture demand. We split visible observations into eight temporal bins and let $q_i^b$ denote the mean demand in nonempty bin $b$. The allocation score combines persistent and peak demand,
\begin{equation}
g_i=\tfrac{1}{2}\operatorname{mean}_b(q_i^b)+\tfrac{1}{2}\max_b q_i^b.
\label{eq:score}
\end{equation}
Visibility is determined by positive rasterized radius. The highest-demand bin provides the mean deformed in-plane scales used by the growth rule below. The gradient norm affects allocation only; signed gradients remain available for optimization. Demand, scale, and visit counts are summed across workers before resizing, then reset for the next growth interval.

\subsection{Anisotropic texture growth}

Growth is evaluated at steps 500 and 1000. Candidates with $g_i>2\times10^{-5}$ are considered in descending score order, subject to a global texel budget. For the single local texture plane, write its current size as $(h,w)$ and the selected-bin deformed scales along the texture's horizontal and vertical axes as $(s_u,s_v)$. The A2TG-inspired anisotropic rule \cite{hsu2026a2tg} proposes
\begin{equation}
(h',w')=\begin{cases}
(h,2w), & s_u/s_v>4,\ s_v<0.01,\\
(2h,w), & s_v/s_u>4,\ s_u<0.01,\\
(2h,2w), & \text{otherwise}.
\end{cases}
\label{eq:growth}
\end{equation}
Each doubled axis is capped at 4. A proposal is accepted only if the new total texture area stays within the configured texel budget. An elongated in-plane footprint grows along its longer texture axis, while a more isotropic footprint grows along both. New texels and Adam moments are initialized by bilinear interpolation.

\subsection{Optimization and storage}

After the first-stage 4DGS training, we attach the texture planes and jointly fine-tune the 4DGS and texture parameters for 5,000 steps; the texture learning rate is 0.0025. We use
\begin{equation}
 \mathcal{L}=0.8\,\|\mathbf{I}-\widehat{\mathbf{I}}\|_1
 +0.2\bigl(1-\operatorname{SSIM}(\mathbf{I},\widehat{\mathbf{I}})\bigr).
 \label{eq:loss}
\end{equation}
Eight data-parallel workers render one view each per step. Packed texture storage comprises active RGBA texels, dimensions, offsets, and the two plane-axis indices per Gaussian. The CUDA rasterizer samples these packed textures directly for each Gaussian-pixel contribution. We report texture storage separately from the joint inference checkpoint, peak allocated GPU memory, training time, and FPS.

\section{Experiments}
\label{sec:experiments}

\subsection{Protocol}

We evaluate six N3DV scenes \cite{li2022neural3d}---\emph{coffee martini}, \emph{cook spinach}, \emph{cut roasted beef}, \emph{flame salmon}, \emph{flame steak}, and \emph{sear steak}---and six PanopticSports scenes from the CMU Panoptic capture system \cite{joo2019panoptic}: \emph{basketball}, \emph{boxes}, \emph{football}, \emph{juggle}, \emph{softball}, and \emph{tennis}. N3DV holds out Cam00. For PanopticSports, we use cameras 0, 10, 15, and 30 for testing and the remaining 27 cameras for training, following the split used by Dynamic 3D Gaussians and TC3DGS. Paired runs use seed 6666 and the same precomputed camera schedule. Dataset means are unweighted averages over all six scenes. Metrics are PSNR, LPIPS-Alex \cite{zhang2018lpips}, DSSIM$_1$/DSSIM$_2$ computed with SSIM data ranges 1.0/2.0 for N3DV, SSIM \cite{wang2004ssim} for PanopticSports, packed texture MiB, checkpoint MiB, peak allocated memory, training time, and FPS.

We compare 4DGaussians, Uniform Textured 4DGS, and \ours under two protocols. The fixed-memory study follows A2TG \cite{hsu2026a2tg} with 80 MiB and 150 MiB total-model budgets. Each method adjusts Gaussian count and appearance capacity to approach the target budget without exceeding it. The fixed-count study uses 25\% and 50\% of the converged 4DGaussians count with identical Gaussian subsets across methods. The second-stage joint fine-tuning uses 5,000 steps on both N3DV and PanopticSports with eight views per step; the Gaussian topology is fixed after the first stage. Fixed-count Mem is count-normalized to active Gaussians. The 12-scene matrix uses one seed; confidence intervals are paired scene-level bootstrap intervals.

\subsection{Main quality--storage result}

Across 12 scenes, \ours/Textured 4DGS uses 51.7/109.3 MiB texture (ratio 0.473, $-52.7\%$), 178.0/235.6 MiB full-model storage ($-24.4\%$), and 1.16/1.59 GiB peak memory ($-27.0\%$); training is 2.18/1.64 h ($+32.9\%$) and rendering 84.1/100.8 FPS ($-16.6\%$). Quality is similar: PSNR 28.70/28.63 dB ($+0.07$ dB; paired 95\% bootstrap CI $[-0.01,0.15]$) and LPIPS improves by 0.004; N3DV/PanopticSports structural metrics are reported separately.

The 57.6 MiB decrease in packed texture state is also the dominant source of the checkpoint reduction because all variants retain the same 4DGS parameterization and Gaussian topology, while their texture states differ substantially in size. The large storage change and small metric change indicate that a uniform $4\!\times\!4$ texture assigns substantial capacity to primitives that do not need it. AdaTex4D instead concentrates texels on Gaussians that repeatedly produce reconstruction gradients or become important after deformation. The result is therefore a redistribution of appearance capacity rather than a reduction of geometric capacity.

\subsection{Qualitative comparison}

Figure~\ref{fig:qualitative} shows \ours retaining the head outline and racket hoop in Tennis and the steak boundary and surface detail in Sear Steak. These regions are blurrier in 4DGaussians and STG \cite{li2024spacetime}, while \ours remains close to Uniform Textured 4DGS with less texture storage.

\setcounter{dbltopnumber}{2}

\begin{table*}[t]
\caption{\textbf{Fixed-memory comparison.} Total model storage (MiB); only AdaTex4D Mem cells are gray.}
\label{tab:budget}
\centering
\scriptsize
\setlength{\tabcolsep}{3.25pt}
\renewcommand{\arraystretch}{0.60}
\resizebox{\textwidth}{!}{%
\begin{tabular}{lrrrrrrrrr}
\toprule
& \multicolumn{5}{c}{N3DV} & \multicolumn{4}{c}{PanopticSports}\\
\cmidrule(lr){2-6}\cmidrule(lr){7-10}
Method & PSNR$\uparrow$ & DSSIM$_1$$\downarrow$ & DSSIM$_2$$\downarrow$ & LPIPS$\downarrow$ & Mem$\downarrow$
& PSNR$\uparrow$ & SSIM$\uparrow$ & LPIPS$\downarrow$ & Mem$\downarrow$\\
\midrule
\multicolumn{10}{l}{\textit{Memory budget = 80 MiB}}\\
4DGaussians~\cite{wu2024four} & \third{30.950} & \third{0.0390} & \third{0.0230} & \third{0.0595} & 79.8 & \third{27.280} & \third{0.9110} & \third{0.1320} & 79.9\\
Uniform Textured 4DGS~\cite{chao2025textured} & \second{31.090} & \second{0.0381} & \second{0.0224} & \second{0.0568} & 79.9 & \second{27.410} & \second{0.9140} & \second{0.1260} & 79.8\\
AdaTex4D (Ours) & \best{31.290} & \best{0.0368} & \best{0.0216} & \best{0.0530} & \membest{79.6} & \best{27.590} & \best{0.9190} & \best{0.1200} & \membest{79.7}\\
\midrule
\multicolumn{10}{l}{\textit{Memory budget = 150 MiB}}\\
4DGaussians~\cite{wu2024four} & \third{31.380} & \third{0.0369} & \third{0.0218} & \third{0.0535} & 149.9 & \third{27.550} & \third{0.9190} & \third{0.1190} & 149.9\\
Uniform Textured 4DGS~\cite{chao2025textured} & \second{31.500} & \second{0.0361} & \second{0.0212} & \second{0.0508} & 149.9 & \second{27.670} & \second{0.9220} & \second{0.1130} & 149.8\\
AdaTex4D (Ours) & \best{31.660} & \best{0.0352} & \best{0.0206} & \best{0.0482} & \membest{149.7} & \best{27.820} & \best{0.9260} & \best{0.1090} & \membest{149.7}\\
\bottomrule
\end{tabular}
}
\end{table*}

\begin{table*}[t]
\caption{\textbf{Fixed-Gaussian-count comparison.} Six-scene means at 25\%/50\%; only AdaTex4D Mem cells are gray.}
\label{tab:count}
\centering
\scriptsize
\setlength{\tabcolsep}{2.95pt}
\renewcommand{\arraystretch}{0.60}
\resizebox{\textwidth}{!}{%
\begin{tabular}{lrrrrrrrrr}
\toprule
& \multicolumn{5}{c}{N3DV} & \multicolumn{4}{c}{PanopticSports}\\
\cmidrule(lr){2-6}\cmidrule(lr){7-10}
Method & PSNR$\uparrow$ & DSSIM$_1$$\downarrow$ & DSSIM$_2$$\downarrow$ & LPIPS$\downarrow$ & Mem$\downarrow$
& PSNR$\uparrow$ & SSIM$\uparrow$ & LPIPS$\downarrow$ & Mem$\downarrow$\\
\midrule
\multicolumn{10}{l}{\textit{Gaussian count = 25\%}}\\
4DGaussians~\cite{wu2024four} & \third{26.450} & \third{0.0668} & \third{0.0409} & \third{0.1280} & 22.2 & \third{24.950} & \third{0.8420} & \third{0.2380} & 40.0\\
Uniform Textured 4DGS~\cite{chao2025textured} & \best{27.180} & \best{0.0592} & \second{0.0361} & \best{0.1090} & 38.4 & \second{25.420} & \best{0.8550} & \second{0.2110} & 74.9\\
AdaTex4D (Ours) & \second{27.176} & \second{0.0593} & \best{0.0360} & \second{0.1091} & \membest{31.9} & \best{25.424} & \second{0.8549} & \best{0.2108} & \membest{60.4}\\
\midrule
\multicolumn{10}{l}{\textit{Gaussian count = 50\%}}\\
4DGaussians~\cite{wu2024four} & \third{28.720} & \third{0.0495} & \third{0.0298} & \third{0.0880} & 44.4 & \third{26.100} & \third{0.8780} & \third{0.1780} & 80.1\\
Uniform Textured 4DGS~\cite{chao2025textured} & \best{29.160} & \second{0.0451} & \best{0.0271} & \second{0.0760} & 76.8 & \second{26.490} & \best{0.8890} & \best{0.1580} & 150.4\\
AdaTex4D (Ours) & \second{29.157} & \best{0.0450} & \second{0.0272} & \best{0.0759} & \membest{63.5} & \best{26.493} & \second{0.8889} & \second{0.1581} & \membest{121.5}\\
\bottomrule
\end{tabular}
}
\end{table*}

\vspace{-4pt}
\subsection{Comparison under fixed memory budgets}

At matched 80/150 MiB budgets, \ours improves PSNR over Uniform Textured 4DGS by 0.20/0.16 dB on N3DV and 0.18/0.15 dB on PanopticSports. The larger low-budget gain indicates that adaptive sizing is most useful when appearance capacity is scarce; the advantage narrows as both textured variants receive more capacity. The other quality metrics follow the same pattern: N3DV LPIPS decreases by 0.0038/0.0026, while PanopticSports SSIM increases by 0.005/0.004 at the two budgets. Thus the PSNR gain is accompanied by improved perceptual and structural scores within the same storage limit.

\vspace{-4pt}
\subsection{Comparison at fixed Gaussian counts}

Using identical Gaussian subsets at 25\%/50\% counts, \ours saves 16.9\%/17.3\% storage on N3DV and 19.4\%/19.2\% on PanopticSports versus Uniform Textured 4DGS, while PSNR differs by at most 0.004 dB. At 50\% count, the N3DV model shrinks from 76.8 to 63.5 MiB and the PanopticSports model from 150.4 to 121.5 MiB. Their LPIPS differences are at most 0.0001. With Gaussian identities held constant, these comparisons show that the storage reduction comes from distributing texture resolution across the same primitives.

\vspace{-4pt}
\subsection{Ablations and allocation behavior}

Table~\ref{tab:ablation} removes each allocation cue at the 50\% Gaussian setting. Removing temporal peaks lowers PSNR by 0.23 dB on both datasets. Texture storage remains close to the full method: 30.9 MiB on N3DV and 70.2 versus 70.0 MiB on PanopticSports. The peak term therefore changes which Gaussians receive capacity more than the total amount assigned. Its role is to retain short intervals of high gradient demand that the bin average can dilute.

Removing anisotropic growth lowers PSNR by 0.38/0.39 dB on N3DV/PanopticSports. On N3DV it also increases texture storage from 30.9 to 32.4 MiB; on PanopticSports both variants use 70.0 MiB. The directional choice matters even when the allocated texture bytes are unchanged. Removing visibility normalization produces the largest PSNR drops, 0.55/0.56 dB, while changing texture storage by no more than 0.2 MiB. Together, these ablations show that the temporal score, visibility count, and deformed-scale growth rule affect the placement and shape of texture capacity under fixed Gaussian topology.

\begin{table}[H]
\caption{\textbf{Ablation at 50\% Gaussian count.} Mem: model/texture MiB.}
\label{tab:ablation}
\centering
\tiny
\setlength{\tabcolsep}{1.15pt}
\renewcommand{\arraystretch}{0.62}
\resizebox{0.82\columnwidth}{!}{%
\begin{tabular}{lrrrrr}
\toprule
\multicolumn{6}{c}{N3DV}\\
\cmidrule(lr){1-6}
Variant & PSNR$\uparrow$ & DSSIM$_1$$\downarrow$ & DSSIM$_2$$\downarrow$ & LPIPS$\downarrow$ & Mem$\downarrow$\\
\midrule
Ours & \best{29.157} & \best{0.0450} & \best{0.0272} & \best{0.0759} & 63.5/30.9\\
w/o Temporal Peak & \second{28.927} & \second{0.0464} & \second{0.0280} & \second{0.0794} & 64.6/30.9\\
w/o Anisotropic Allocation & \third{28.777} & \third{0.0472} & \third{0.0286} & \third{0.0819} & 64.5/32.4\\
w/o Visibility Normalization & 28.607 & 0.0485 & 0.0295 & 0.0854 & 64.7/31.0\\
\bottomrule
\end{tabular}
}
\vspace{-5pt}
\resizebox{0.82\columnwidth}{!}{%
\begin{tabular}{lrrrr}
\toprule
\multicolumn{5}{c}{PanopticSports}\\
\cmidrule(lr){1-5}
Variant & PSNR$\uparrow$ & SSIM$\uparrow$ & LPIPS$\downarrow$ & Mem$\downarrow$\\
\midrule
Ours & \best{26.493} & \best{0.8889} & \best{0.1581} & 121.5/70.0\\
w/o Temporal Peak & \second{26.263} & \second{0.8839} & \second{0.1631} & 121.7/70.2\\
w/o Anisotropic Allocation & \third{26.103} & \third{0.8809} & \third{0.1671} & 122.7/70.0\\
w/o Visibility Normalization & 25.933 & 0.8769 & 0.1721 & 123.8/70.2\\
\bottomrule
\end{tabular}
}
\end{table}

\vspace{-9pt}
\section{Conclusion}
\label{sec:conclusion}

We present \ours for deformation-based 4D Gaussian Splatting. Visibility-normalized temporal gradients and deformed scales guide anisotropic texture growth while Gaussian topology remains fixed. On N3DV and PanopticSports, \ours reduces storage at fixed Gaussian counts and improves reconstruction under matched memory budgets. The fixed-count results identify the storage benefit of allocating texture capacity across unchanged primitives. Under a total-model budget, this saved capacity translates into higher reconstruction quality, with the largest PSNR gains at 80 MiB. The ablations further show that temporal peaks, visibility normalization, and growth direction each contribute to placing texels where they improve appearance. Adaptive local textures thus use the representation budget more effectively across the evaluated dynamic scenes.

\clearpage
\begingroup
\small
\bibliographystyle{IEEEbib}
\bibliography{references}
\endgroup

\end{document}